\pdfoutput=1
\documentclass[11pt]{article}

\usepackage[]{naacl2021}

\usepackage{times}
\usepackage{latexsym}
\usepackage{hyperref}       % hyperlinks
\usepackage{url}            % simple URL typesetting
\usepackage{booktabs}       % professional-quality tables
\usepackage{amsfonts}       % blackboard math symbols
\usepackage{nicefrac}       % compact symbols for 1/2, etc.
\usepackage{microtype}      % microtypography
\usepackage{graphicx}
\usepackage{float}
\usepackage{hhline}
\usepackage{bbm}
\usepackage{caption}
\usepackage{amsmath}
\usepackage{tabularx}
\usepackage{multirow}
\usepackage{mathrsfs}
\usepackage[utf8]{inputenc} % allow utf-8 input
\usepackage[T1]{fontenc}    % use 8-bit T1 fonts

\usepackage[T1]{fontenc}
\usepackage[utf8]{inputenc}

\usepackage{microtype}

\title{A First Look: Towards Explainable TextVQA Models via Visual and Textual Explanations}

 \author{Varun Nagaraj Rao  \textsuperscript{* $\dagger$}, Xingjian Zhen \textsuperscript{* $\ddagger$ $\mathsection$ } , Karen Hovsepian \textsuperscript{$\dagger$} , Mingwei Shen \textsuperscript{$\dagger$}  \\
         \textsuperscript{$\dagger$} PARS \textsuperscript{$\P$}, Amazon.com, Seattle \ \ \ \ \ \textsuperscript{$\ddagger$} University of Wisconsin-Madison \\ 
 \begin{small}
 	\texttt{varao@amazon.com, xzhen3@wisc.edu, \{khhovsep, mingweis\}@amazon.com}
 \end{small} \\
\begin{small} \textsuperscript{*} Equal Contribution \ \ \ \ \textsuperscript{$\mathsection$} Work done while an intern at Amazon \end{small} }
\begin{document}
\maketitle

\begin{abstract}
Explainable deep learning models are advantageous in many situations. Prior work mostly provide unimodal explanations through post-hoc approaches not part of the original system design. Explanation mechanisms also ignore useful textual information present in images. In this paper, we propose MTXNet, an end-to-end trainable multimodal architecture to generate multimodal explanations, which focuses on the text in the image. We curate a novel dataset TextVQA-X, containing ground truth visual and multi-reference textual explanations that can be leveraged during both training and evaluation. We then quantitatively show that training with multimodal explanations complements model performance and surpasses unimodal baselines by up to 7\% in CIDEr scores and 2\% in IoU. More importantly, we demonstrate that the multimodal explanations are consistent with human interpretations, help justify the models' decision, and provide useful insights to help diagnose an incorrect prediction. Finally, we describe a real-world e-commerce application for using the generated multimodal explanations.
\renewcommand*{\thefootnote}{\fnsymbol{footnote}}
\footnotetext{\textsuperscript{$\P$} Product Assurance, Risk, and Security \url{https://www.amazon.jobs/en/teams/product-assurance-risk-security}}
\renewcommand*{\thefootnote}{\arabic{footnote}}
 %\footnotetext[0]{\textsuperscript{$\P$}https://www.amazon.jobs/en/teams/product-assurance-risk-security}
\end{abstract}

\section{Introduction}

%para1: general introduction on explainability, justifying decisions
%para2: specific explainable textvqa task with previous approaches citations.
%para3: limitations of previous work
%para4: "in this paper" we propose etc....
%para5: bullet points for contributions

The ability to explain decisions through voice, text and visual pointing, is inherently human. Deep learning models on the other hand, are rather opaque black boxes that don't reveal very much about how they arrived at a specific prediction. Recent research effort, aided by regulatory provisions such as GDPRs ``right to explanation'' \cite{goodman2017european}, have focused on peeking beneath the hood of these black boxes and designing systems that inherently enable explanation. Explainable multimodal architectures can also be used to reduce the effort required for manual compliance checks of products sold by online retailers. Further, explanations can be provided as evidence to justify decisions and help improve customer and seller partner experiences. 

We choose the TextVQA task proposed by \citet{singh2019towards} for realizing the system, motivated by two reasons. First, the task is multimodal and is naturally suited for generating multimodal explanations. Second, the task specifically focuses on the text in the image, known to encode essential information for scene understanding and reasoning \cite{hu2020iterative}, and allows for better quality of explanations including the text recognized. Several approaches have been proposed for the TextVQA task \cite{singh2019towards, hu2020iterative, mishra2019ocr, biten2019scene, kant2020spatially}, but they do not include a means for explaining the model decision. In addition to allowing humans to interpret the model's decision, we believe the explanations can also provide valuable insight into what component could be improved.

\begin{figure}[tb]
	\centering
	\includegraphics[width=1\linewidth]{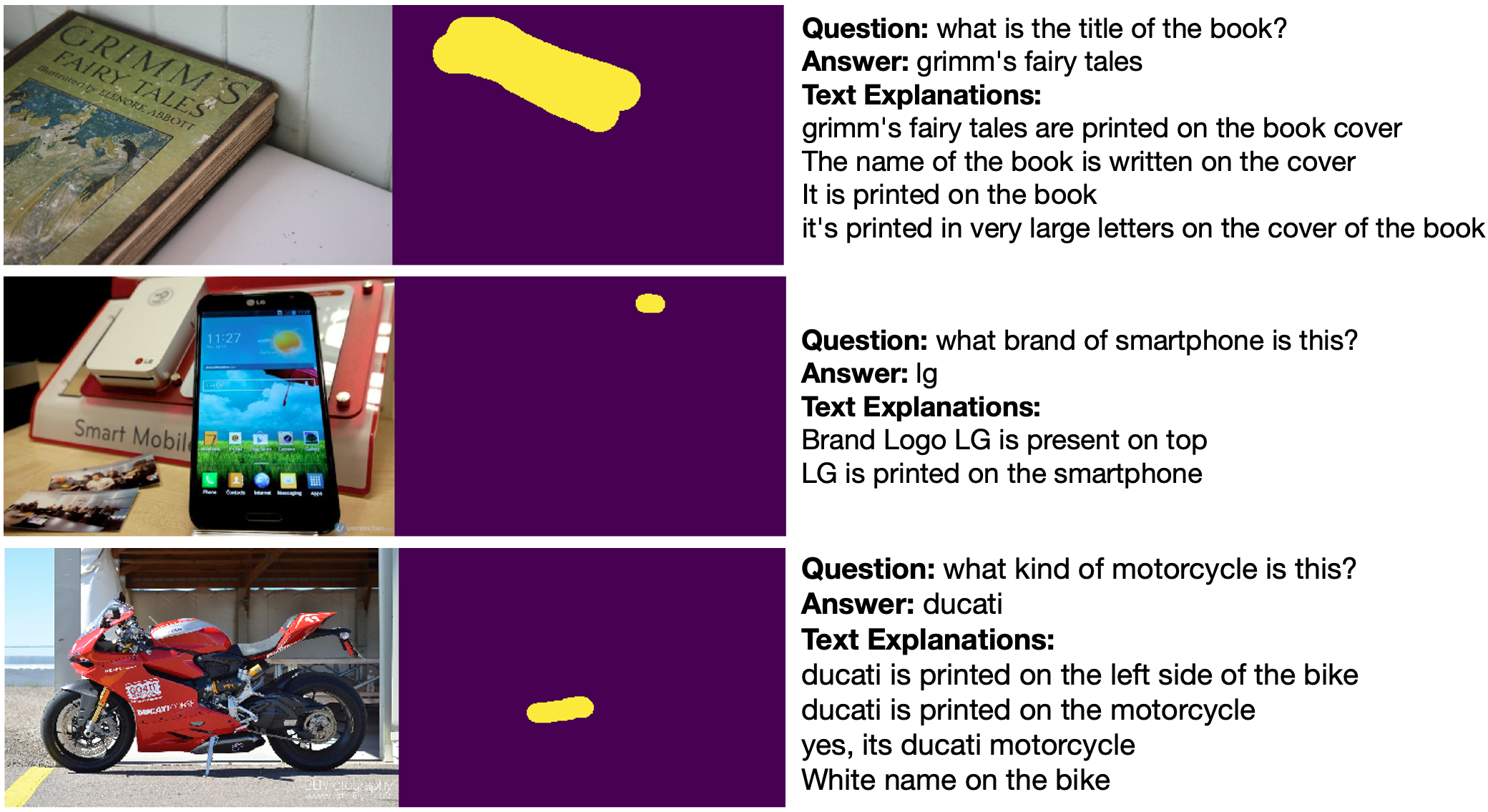}
	\caption{Sample Ground Truth Labels}
	\label{fig:annotationsamples}
	\vspace{-1em}	
\end{figure}

Most prior explanation approaches \cite{hendricks2016generating, anne2018grounding, li2018tell} have been unimodal and do not focus on the text in the image.  Only recently, \citet{huk2018multimodal} and \citet{wu2019faithful} generated multimodal explanations for the VQA and Activity Recognition tasks.  They curated datasets (VQA-X, ACT-X) consisting of single reference ground truth textual explanations and relied on implicit attention-based visual explanations without any access to labeled visual ground truth. However, their models cannot read and incorporate text in the image into the explanations. In addition, it is debatable whether attention mechanisms are indeed explanations \cite{wiegreffe2019attention, jain2019attention}. Moreover, other works \cite{das2017human} have shown that current VQA attention models do not seem to look at the same regions as humans, resulting in inconsistent explanations. 

The goal of our work is two-fold. First, to collect a multimodal explanations dataset (TextVQA-X) thereby highlighting the need to curate datasets where explanations are not post-hoc but part of the initial interpretable model design. Non post-hoc explanations which may not be faithful to the model decision but are in line with human explanations are still beneficial to end users. Figure \ref{fig:annotationsamples} provides a representative example. Second, to implement a multimodal explanation system that has the ability to not only read and reason about the text in the image, but more importantly justify its decision with natural language and visually highlight the evidence, useful to even non-experts \cite{miller2017explainable}. The explanations and model decision must be tightly coupled and mutually influence each other through an end-to-end trainable architecture. In summary, our contributions are as follows: \vspace{-0.45em}

\begin{itemize}
	\itemsep0em 
	\parskip0em
	\item We present TextVQA-X, a novel dataset of human-annotated multimodal explanations that includes ground truth segmentation maps and multi-reference textual explanations containing text in the image. The raw dataset is available publicly \footnote{\url{https://github.com/amzn/explainable-text-vqa}}. (Section \ref{sec:textvqax})
	\item We propose the first end-to-end trainable MTXNet architecture that produces high quality textual and visual explanations, focusing on the text in the image.  (Section \ref{sec:mtxnet})
	\item Qualitative and quantitative results show that textual and visual explanations help justify a model's decision and help diagnose the reasons for an incorrect prediction. (Section \ref{sec:expts})
	\item We describe a real-world e-commerce system that can leverage the multimodal explanations and also highlight its challenges. (Section \ref{sec:application})
\end{itemize}

\section{Related Work}

\noindent \textbf{VQA / TextVQA.} The VQA task \cite{antol2015vqa} has received a lot of research attention in terms of both datasets \cite{antol2015vqa,johnson2017clevr, hudson2019gqa} and methods \cite{,anderson2018bottom, ben2017mutan, lu2019vilbert}. Oftentimes however, these models predict an answer without completely understanding the question and do not change answers across images \cite{agrawal2016analyzing}. Further, they ignore the text in the image and tend to focus on visual components such as objects. To address this limitation, the TextVQA task was proposed by \citet{singh2019towards} and has received recent research attention \cite{kant2020spatially, hu2020iterative, biten2019scene, mishra2019ocr}.  However, not having reliable explanation mechanisms that focus on the text in the image, as part of the system design makes it difficult to diagnose prediction failures. Our work, thus allows for better diagnosis of model failures through explanations in line with human interpretations and focus on the text in the image.

\noindent \textbf{Explanations.} Prior explanation approaches \cite{shortliffe1975model, van2004explainable, zeiler2014visualizing, goyal2016towards, ribeiro2016should, selvaraju2017grad, das2017human} focus on parts of the input that is relevant to the model's decision, but not on explicitly generating explanations as model predictions. \citet{hendricks2016generating, anne2018grounding} were the first to generate natural language justifications for image classifiers. Unlike our model however, explanations are unimodal and there are no reference human explanations. Closer to our objective \citet{huk2018multimodal} generate multimodal explanations and curate a new VQA-X dataset. \citet{wu2019faithful} extend their work to ensure explanations can be traced back to an object ensuring local faithfulness. However, their explanations do not contain the text in the image. They use implicit attention for visual explanations and have no access to visual ground truth during training. Further, they use a single textual explanation reference during training. In contrast, our work incorporates multimodal explanations which focuses on the text in the image.

% \noindent \textbf{Graph Neural Networks.} Due to the inner relationship within objects and between the OCR tokens in the images, it would be natural to build a scene graph from the images. Thus, Graph Neural Network could be a good enhancement for the visual feature embedding which can encode these relationships. \citet{kipf2016semi} introduced the Graph Convolutional Networks (GCN) which pass information between neighbors in the graph. By converting the adjacency matrix into Fourier domain, \citet{kipf2016semi} defined the convolution operator and filters for the graphical data. \citet{velivckovic2017graph} introduced self-attention in the graph model to attend to the more relevant neighbors when passing information through the graph. Though Graph Neural Networks are successful in several areas including molecular fingerprints \cite{duvenaud2015convolutional} and social networks \cite{fan2019graph}, they are relatively unexplored in the VQA domain \cite{gao2020structured}.

\section{TextVQA-X Dataset} \label{sec:textvqax}

%\begin{figure*}[htb]
%	\centering
%	\begin{minipage}{0.50\textwidth}
%		\centering
%		\includegraphics[width=0.9\linewidth]{labeling_ui}
%		\captionof{figure}{Sagemaker Ground Truth Labeling UI}
%		\label{fig:labelingui}
%	\end{minipage}
%	\hfill
%	\begin{minipage}{0.49\textwidth}
%		\centering
%		\centering
%		\includegraphics[width=0.9\linewidth]{annotation_samples}
%		\captionof{figure}{Sample Ground Truth Labels}
%		\label{fig:annotationsamples}
%	\end{minipage}
%\end{figure*} 

%\begin{figure}[htb]
%		\centering
%		\includegraphics[width=1\linewidth]{labeling_ui}
%		\caption{Sagemaker Ground Truth Labeling UI}
%		\label{fig:labelingui}
%\end{figure}

To train and evaluate multimodal explanation models that focus on the text in the image, we collect the TextVQA-X dataset by human annotation of a subset of samples from the TextVQA dataset \cite{singh2019towards}.

\begin{figure*}[htb]
	\centering
	\includegraphics[width=\textwidth]{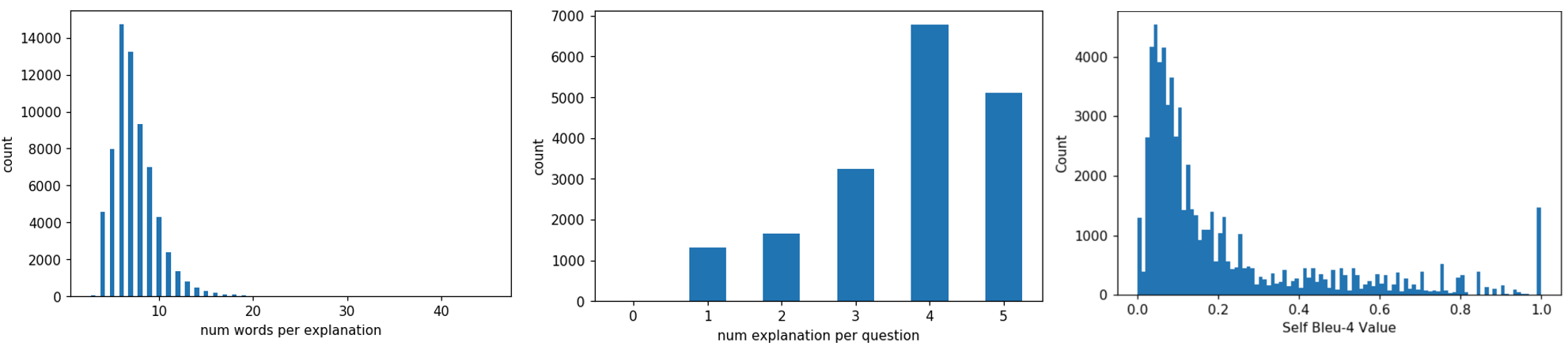}
	\caption{TextVQA-X Dataset Statistics}
	\label{fig:datasetstatistics}
\end{figure*}

\subsection{Ground Truth Label Collection} 
%TODO:

We used the Sagemaker Ground Truth \cite{groundtruth} platform to create a labeling task for gathering visual and textual explanations. Human annotators were asked to provide a single textual explanation that answers the question "Why do you think <answer> is the correct answer for the given question and image pair?". Specific instructions added that annotators should try to incorporate the answer and/or the text in the image as part of their explanation. The annotators were also asked to make use of a brush to segment image regions relevant to both the answer and written explanation. Sample annotations are shown in Figure \ref{fig:annotationsamples}. Each image and question pair can have up to 5 distinct human annotators allowing for multi-reference training and evaluation \cite{zheng2018multi}. A single segmentation map is obtained by using a threshold of 0.5 obtained as an average over all annotations. Bad actors were identified and most were removed through a combination of heuristics and manual checks. Overall, we collected more than 67K explanations among over 800 unique workers.

%challenges of clean annotations, gaming system, spelling errors

\subsection{TextVQA Explanation Dataset (TextVQA-X). }

\begin{table}[htb]
	\small
	\centering
	\begin{tabular}{l|c}
		\hline
		\multicolumn{1}{c|}{\textbf{Dataset Statistic}} & \textbf{Value} \\ \hhline{==}
		Num. Unique Images                                  & 11681          \\
		Num. Questions                                      & 18096          \\
		Num. Unique Questions                               & 15374          \\
		Num. Visual Explanations                            & 67055          \\
		Num. Textual Explanations                           & 67055          \\
		Num. Unique Textual Explanations                    & 61999          \\
		Avg. Num Textual Explanations per Question                      & 3.71           \\
		Avg. Words per Textual Explanation                                   & 7.36           \\
		Avg. Characters per Textual Explanation                          & 36.92          \\
		Textual Explanation Vocab Size                 & 17910         \\ \hhline{==}
	\end{tabular}
	\caption{TextVQA-X Dataset Summary}
	\label{tab: textvqa-x_summary}
\end{table}

%TODO talk about self blue-1

%Distributional statistics of the explanations are present in Figure \ref{fig:datasetstatistics}. 
In order to obtain a measure of the quality of explanations and to help filter out bad actors, we make use of the Self-BLEU-4 metric \cite{zhu2018texygen}. The Self-BLEU score is used to measure how one sentence resembles the rest in a generated collection by regarding one sentence as the hypothesis and the rest as references. A higher Self-BLEU score implies higher similarity of the hypothesis with all the references. A lower Self-BLEU implies higher diversity and lesser overlap. Although we would like to have several diverse textual explanations, we noticed that most good textual explanation annotations have overlap with others. The average Self-BLEU-4 across all annotations was 0.21 indicating consistent overlap and quality. 

\noindent \textbf{Comparison with VQA-X and VQA-HAT datasets}. With respect to textual explanations, the TextVQA-X includes multi-references with an average of 3.71 explanations for each QA pair that can be utilized for both training and testing. In contrast, VQA-X \cite{huk2018multimodal} contains an average of 1.27 explanations with a single textual explanation for QA pairs in the training set and three textual explanations for test/val QA pairs. VQA-HAT \cite{das2017human} does not include textual explanations. As far as visual explanations are concerned, there are a number of distinctions among these datasets. First, both VQA-X and VQA-HAT are defined on the VQA task, which does not require reading text in the. In contrast, the TextVQA-X is specifically designed to focus on the text in the image. Second, TextVQA-X includes one ground truth visual explanation for both training and testing (total 67K), whereas VQA-X includes explanations only as part of testing for a small random subset (total 6K). And third, similar to VQA-X, TextVQA-X annotators were asked to directly segment the relevant image region. On the contrary, VQA-HAT annotations were collected by having humans unblur the images and are more likely to introduce noise when irrelevant regions are uncovered.

%TODO:
% (1) table with dataset statistics (maybe also have comparison with VQA-X) 
% (2) image num words per explanation
% (3) image num expl per question
% (4) image self bleu4 distribution

\section{Multimodal Text-in-Image Explanation Network (MTXNet)} \label{sec:mtxnet}

\begin{figure*}[htb]
	\centering
	\includegraphics[width=0.8\linewidth, height=6cm]{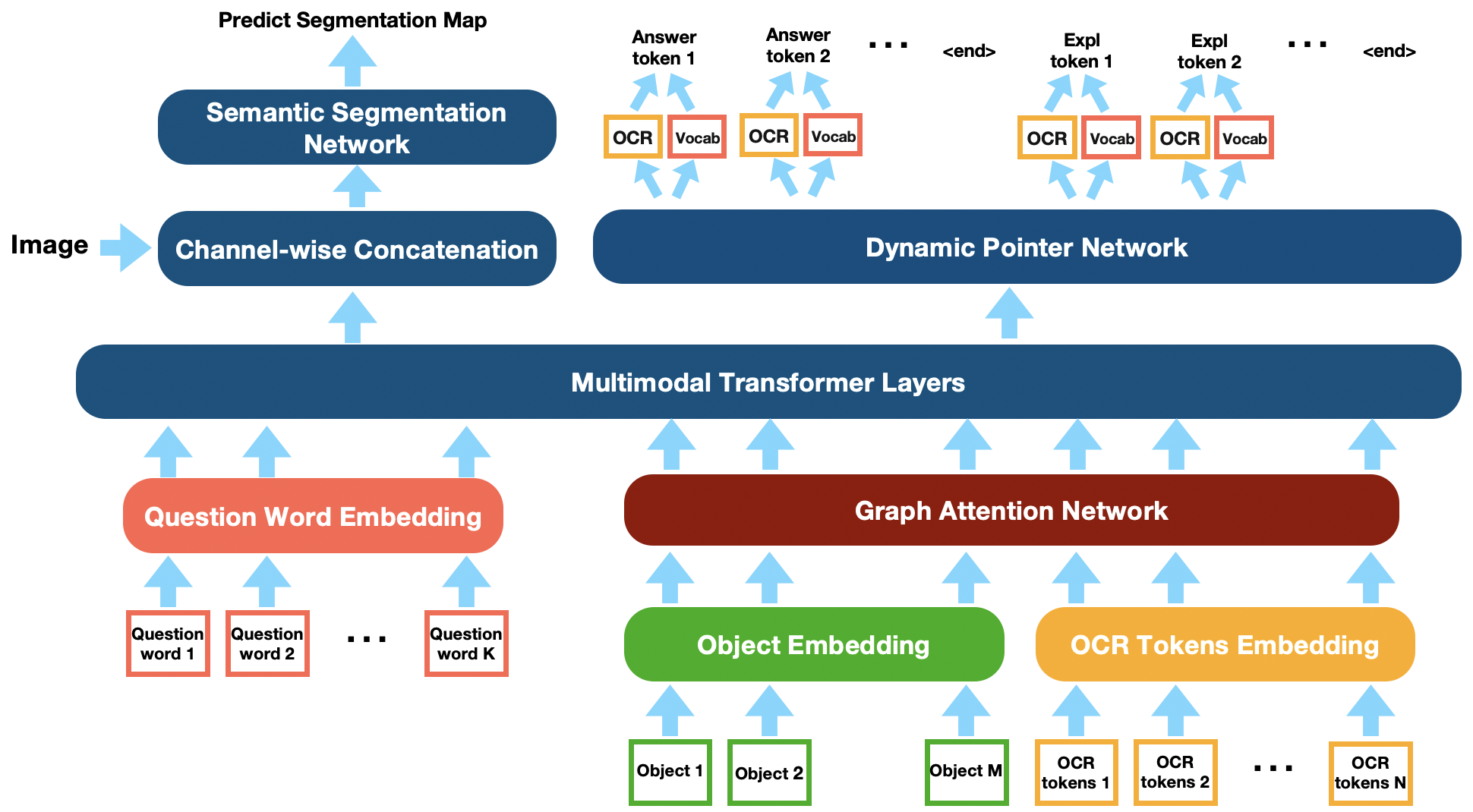}
	\caption{Our Multimodal Text-in-Image Explanation Model (MTXNet) architecture generates multimodal explanations. Explanations and Answers are utilized as a part of the iterative autoregressive decoding procedure.}
	\label{fig:model}
\end{figure*}

%TODO: Brief Introduction about the model
% image overall model architecture

%The goal of this work is two-fold. First, to collect a multimodal explanations dataset thereby highlighting the need to curate datasets where explanations are not post-hoc but part of the initial interpretable model design. And second, to implement a multimodal explanation system that has the ability to not only read and reason about the text in the image, but more importantly justify its decision with natural language and visually highlighting the evidence. Moreover, the explanations and model decision must be tightly coupled and mutually influence each other through an end-to-end trainable architecture.

We design our Multimodal Text-in-Image Explanation Network (MTXNet) to allow for end-to-end multitask training of answer prediction, text generation and semantic segmentation extending the M4C model proposed in \cite{hu2020iterative}. In the subsequent subsections we describe each of the individual components in more detail.

\subsection{Graph Attention Network (GAT)}

\begin{figure}[H]
	\centering
	\includegraphics[width=0.8\linewidth]{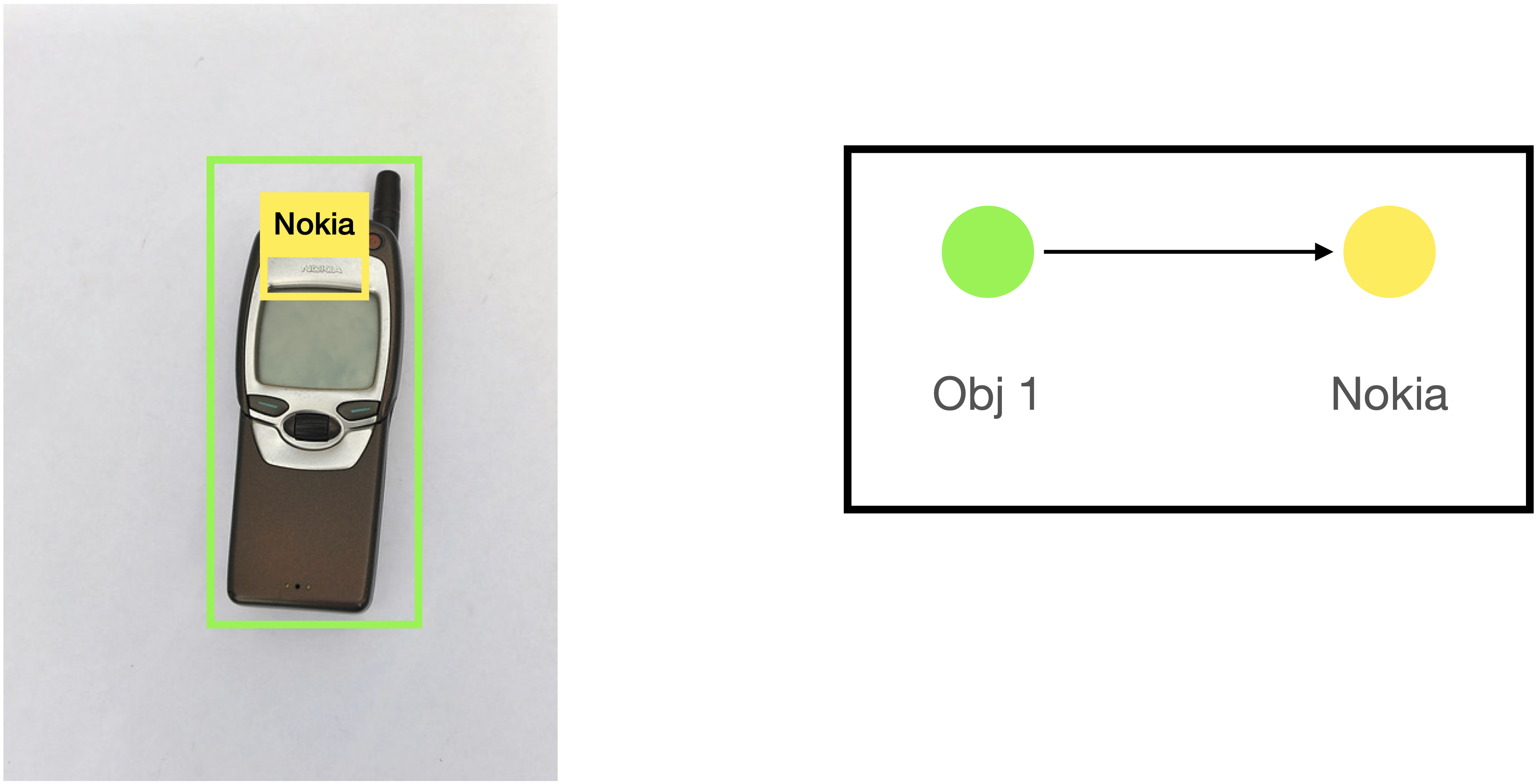}
	\caption{An example of how to build the graph}
	\label{fig:buildgraph}
\end{figure}
Many questions in the TextVQA dataset require the model to acknowledge the spatial relationship between objects and OCR tokens. To better encode the relationship between objects and OCR tokens and subsequently generate better quality explanations, we leverage graph neural networks.  The ideal way to build the graph is to link together relevant components such as question words, OCR tokens and object labels. However, there are two limitations in the existing TextVQA dataset that prevent us from adopting this approach. First, the OCR tokens may be misspelled due to an inaccurate OCR system. And second, the object labels are not included and only the bounding box coordinates are present. Thus, for our model we build the graph using only the visual inputs (object and OCR region bounding boxes). Each object location and OCR token is treated as a node in the graph. Whenever the bounding box associated with node \textit{i} is contained in node \textit{j}, we add an edge from node \textit{j} to node \textit{i}. An example is presented in Figure \ref{fig:buildgraph}.  We then make use of the Graph Attention Network (GAT) \cite{velivckovic2017graph} to operate on the structured data. Unlike Graph Convolutional Networks (GCN) \cite{kipf2016semi} that treat each adjacent node equally, GATs incorporate attention into the layer-wise propagation rule and allows the model to variably weigh adjacent nodes based on relevancy.  

%TODO write that GAT better suited to improve textual explanation
\subsection{Multimodal Transformer (MMT)}
The multimodal transformer operates on three modalities - question words, visual objects and OCR tokens. The feature definitions are identical to that proposed in M4C \cite{hu2020iterative} with the addition of textual explanation embeddings whose embedding process resembles that of the question words. The object embedding is obtained as a combination of the 2048-dim Faster R-CNN detector output and 4-dimensional relative location feature $[x_{min}/W_{im}, y_{min}/H_{im}, x_{max}/W_{im}, y_{max}/H_{im}]$. The OCR token embedding is obtained as a combination of 300-dim FastText vector \cite{bojanowski2017enriching}, 2048-dim output from fc6 features/ fc7 weights from Faster R-CNN detector for the bounding box region, 604-dim Pyramidal Histogram of Characters (PHOC) vector \cite{almazan2014word}, and 4-dim relative location feature $[x_{min}/W_{im}, y_{min}/H_{im}, x_{max}/W_{im}, y_{max}/H_{im}]$. Features are projected to a common \textit{d}-dimensional semantic space used for decoding and prediction. The prediction takes place through a dynamic pointer network \cite{vinyals2015pointer} that allows to either predict from a fixed vocabulary or from OCR tokens extracted from the image.

%TODO write about why prefix LM is important here.

\subsection{Multireferences for Textual Explanations}
Neural text generation tasks such as machine translation, image captioning and summarization typically only consider a single reference for each example during training \cite{zheng2018multi}. In our case however, considering just a single reference for training is insufficient because of the inherently subjective nature of textual explanations.  Thus we leverage the multi-references we have collected in the TextVQA-X dataset during both training and evaluation. We use the \textit{sample one} technique for incorporating multi-references during training. We randomly pick one of the available references in each training epoch.

%\subsection{Cross Modal Feedback} 
%In \cite{huk2018multimodal} the explanations are generated by conditioning on the image, question and answer features and decoding using an LSTM. The dependency is thus unidirectional, with the answer influencing the explanation but explanations not directly influencing the answer. In contrast, MTX-Net incorporates bi-directional dependencies between the answer and explanations through the MMT during forward pass and end-to-end backpropagation during the backward pass. 
%
%To incorporate this cross modal feedback, we propose a randomised 3-phase training of the multi-head self attention mechanism. In the first phase, the answer is generated before the explanation. When generating the explanation, the model can attend to not only the explanation tokens before the current location, but also all the answer tokens generated. However, if we always train in this order, an imbalance between the answer and the textual explanation modalities will be introduced since the explanation depends on the answer. To counter this, in the second phase we generate the textual explanation first and then generate the answer allowing the model to attend to all the explanation words generated before  the answer prediction. Finally, in the third phase, we restrict the freedom for cross modal interaction and allow the model to only attend to tokens within its modality to reinforce intra-modal interactions.

\subsection{Visual Explanations through Semantic Segmentation}
Visual explanations are obtained through a semantic segmentation module (Feature Pyramid Network - FPN \cite{kirillov2017unified}). They are made an explicit and natural component of end-to-end training by leveraging ground truth label supervision. Incorporating explicit visual explanations is known to achieve state-of-the-art results on semantic segmentation benchmarks \cite{li2018tell}. Moreover, this allows the model to explain the image region in focus, while also providing a means for feedback.  On another note, in the complimentary domain of NLP, the use of attention as a means of model explanation has been a topic of considerable debate \cite{wiegreffe2019attention, jain2019attention}. We thus leverage ground truth label supervision and explicitly ensure the visual explanation to be part of the training objective.  To incorporate the multimodal embedding from the MMT into the segmentation module, we reshape, pad and concatenate the output with the raw input image along the channel. Thus, the overall input channels for the segmentation module increases to five, with 3 color channels and 2 multimodal channels. The output of the segmentation model is a continuous mask with a higher value implying greater relevancy to the inputs. The mask may be binarized through thresholding.

\subsection{Training}
%TODO talk about loss function and learnable parameters
The MTXNet architecture is end-to-end trainable with three distinct tasks (1) answer prediction (2) textual explanation generation and (3) visual explanation through semantic segmentation. We ensure cross-modal feedback between the textual explanations and predicted answers by leveraging a phased training process where we randomly choose between one of three choices (1) predict answer then textual explanation (2) predict textual explanation then answer and (3) predict both answer and textual explanation independently. Each task corresponds to an individual part of the training objective.  For the losses of answer prediction ($\mathscr{L}_{\text {ans}}$) and textual explanation generation ($\mathscr{L}_{\text {text}}$) we use the \textit{binary cross entropy with logits} \footnote{\url{https://pytorch.org/docs/stable/generated/torch.nn.BCEWithLogitsLoss.html}}. For semantic segmentation ($\mathscr{L}_{\text {vis}}$) we use the \textit{dice loss} \cite{sudre2017generalised}.  The naive approach to combine multiple losses is to use a predetermined weighted linear sum of the individual losses. However, the model performance is sensitive to the weights which are hyperparameters and expensive to tune. We thus use a multitask learning loss with homoscedastic uncertainty as proposed by \citet{kendall2018multi}. The overall objective is present in Equation \ref{eq:multitaskloss}. The weights \{$w_{ans}, w_{text}, w_{vis}$\} corresponding to the loss terms of the three individual tasks are learned. 
\begin{align}	\label{eq:multitaskloss}
	\small
	\begin{split}	
		\mathscr{L} = \sum_{i} \mathscr{L}_i exp(-w_i) + w_i, i \in \{ans,text,vis\}
	\end{split}
\end{align}

\section{Experiments} \label{sec:expts}
In this section, we detail the experimental setup, present quantitative results with ablations and finally analyze qualitative results.

\subsection{Experimental Setup}
This subsection discusses the dataset splits, model training, hyperparameter settings and evaluation metrics.

\noindent \textbf{Dataset Splits.} We use the TextVQA-X dataset described in Section \ref{sec:textvqax}. We choose a random 80/20 split for train and test. The dataset split statistics are present in Table \ref{tab:datasplits}. Each question is associated with a single image, one or more textual explanations and a single visual explanation. The OCR tokens and object regions are already present in the original TextVQA dataset.

\begin{table}[htb]
	\small
	\centering
	\begin{tabular}{c|c|c|c|c}
		\hline
		\textbf{Split}& \textbf{\#Img.}	&  \textbf{\#Ques.} & \textbf{\#Text Expl.} & \textbf{\#Vis. Expl.} \\ \hhline{=|=|=|=|=}
		train & 10379 & 14475 & 53536 & 14475 \\
		test & 3354 & 3619 &13507 & 3619 \\ \hline
	\end{tabular}
	\caption{Train / Test Splits of TextVQA-X Dataset}
	\label{tab:datasplits}
	
\end{table}

\noindent \textbf{Preprocessing.} The dynamic pointer network is allowed to choose between a fixed 5000 word vocabulary and a maximum of 100 OCR tokens per image. For each image, we use the top 36 possible objects extracted by Faster R-CNN sorted in descending order of confidence score attribute. The average number of edges per image was 104. Each image included an average of 13 OCR tokens. The text explanations and answers are capped to a maximum length of 16 and 12 tokens respectively. For the visual explanations, we use a FPN decoder with ResNeXt50 encoder and 320 $\times$ 320 $\times$ 5 input feature size. The MMT consists of 4 layers and 12 attention heads. The dimension of the joint embedding space is $184 \times 768$ which is padded and resized to $320 \times 320 \times 2$ and concatenated with the 3-channel image input.

\noindent \textbf{Model training and hyperparameters.}
%TODO
% mention multireference training random selection
% mention self blue-4 upper bound
We train the MTXNet model end-to-end in a supervised setting using the Pythia \footnote{\url{https://github.com/facebookresearch/mmf}} framework. We use a batch size of 128 and train for a maximum of 8500 epochs using Adam optimizer. The learning rate is set to $1e-4$ with no weight decay. The best model is chosen corresponding to the lowest train loss at an evaluation granularity of every 100 epochs. The entire training task varies from 14-20 hours on 8 Nvidia K80 GPUs. 

\noindent \textbf{Evaluation Metrics.}
Each question in the TextVQA dataset has 10 human-annotated answers, and the predicted answer accuracy is measured via a soft voting in accordance with the VQA task evaluation script \footnote{\url{https://visualqa.org/evaluation}}. We evaluate the textual explanations using the standard BLEU-4 \cite{papineni2002bleu}, ROUGE \cite{lin2004rouge}, METEOR \cite{banerjee2005meteor} and CIDEr \cite{vedantam2015cider} metrics computed with the \texttt{coco-caption} \footnote{\url{https://github.com/tylin/coco-caption}} code . All the text generation metrics account for multi-references by averaging the individual scores. Finally, we evaluate the visual explanations using IoU (Intersection over Union) score with a threshold of 0.5. 

\subsection{Ablation Study}
We ablate MTXNet and compare quantitatively with a related model on our TextVQA-X dataset through automatic evaluations for answers and explanations. The results are present in Table \ref{tab:results}. 

% \usepackage{multirow}
\begin{table*}[htb]
	\small
	\begin{tabular}{l|l|c|cccc}
		\hhline{=======}
		\multicolumn{1}{c|}{\multirow{2}{*}{\textbf{Ablation}}} & \multicolumn{1}{c|}{\multirow{2}{*}{\textbf{Approach}}} &  \multicolumn{1}{l|}{\textbf{Visual Explanation}} & \multicolumn{4}{c}{\textbf{Textual Explanation}}              \\ \cline{3-7} 
		\multicolumn{1}{c|}{}                                   & \multicolumn{1}{c|}{}                                                       & \textbf{IoU}                                     & \textbf{B}    & \textbf{R}    & \textbf{M}    & \textbf{C}    \\ \hhline{=======}
		No visual explanation  (VE)                             & MTXNet (GAT + MR + TE )                                                       & -                                                & 25.16          & 47.63          & 21.76          & 88.43          \\
		No textual explanation (TE)                             & MTXNet (GAT + MR + VE )                                                                 & 16.10                                             & -             & -             & -             & -             \\ \hline
		No graph attention (GAT)                                & MTXNet (MR + TE + VE )                                                                  & 16.55                                             & 27.87          & 49.28          & 21.61          & 88.57          \\ \hline
		No multireferences (MR)                                 & MTXNet (GAT + TE + VE )                                                                   &                        17.52                         &         5.92      &        28.05       &      11.65         &        70.60       \\ \hline
		Consolidated architecture                               & MTXNet (GAT + MR + TE + VE )                                                    & \textbf{18.86}                                    & \textbf{31.07} & \textbf{53.87} & \textbf{22.06} & \textbf{95.07} \\ \hhline{=======}
	\end{tabular}
	\caption{Quantitative Evaluation of Answer and Explanations. All metrics are in \%. VE: visual explanation, TE: textual explanation, GAT: graph attention network, MR: multi-references. Evaluated automatic metrics: Intersection over Union (IoU), BLEU-4 (B), METEOR (M), ROUGE (R), CIDEr (C).}
	\label{tab:results}
	\vspace{-1em}
\end{table*}

\noindent \textbf{Comparison with existing baselines.} We compute the performance of the baseline model M4C \cite{hu2020iterative} on the TextVQA-X test set (without explanations) and obtain an answer accuracy of 35.23\%. Using the MTXNet architecture and evaluating on the TextVQA-X test set, we obtain an answer accuracy of 36.27\%. The addition of explanations thus complements the MTXNet performance. 

\noindent \textbf{Unimodal vs. Multimodal explanations} We notice that each modality mutually influences the other as the model learns to jointly optimize for both modalities of explanations and the answer prediction. Excluding visual explanations results in the largest drop of up to 7\% in CIDEr scores of the textual explanations. Similarly, the absence of text explanations results in a 2\% drop in IoU of visual explanations. More importantly, we notice that the multimodal explanations provide visual and textual rationale into a models decision. This further accentuates the value of designing multimodal explanation systems.\\
\noindent \textbf{GAT better captures structural dependencies.} The removal of GAT from the MTXNet architecture adversely impacts the quality of explanations and answers. The greatest drop of 7\% is observed for the CIDEr metric. We believe the GAT helps better encode the relationship between objects and OCR tokens enhancing the relationship reasoning ability. The image region corresponding to the text is also highlighted better as seen in the 2\% increase in IoU when GAT is included in MTXNet.  \\
\noindent \textbf{Multi-reference training improves text generation.} Training with multi-references significantly outperforms training with a single randomly chosen sample fixed for all epochs. The largest increase of up to 25\% was noticed in CIDEr score, with the increase being consistent across all text generation metrics. This underscores the benefits of having multi-references for both training and evaluation and designing systems that utilize this effectively.

%TODO overfitting, training accuracy much higher than val

\subsection{Qualitative Samples}

\begin{figure}[htb]
	\centering
	\includegraphics[width=\linewidth]{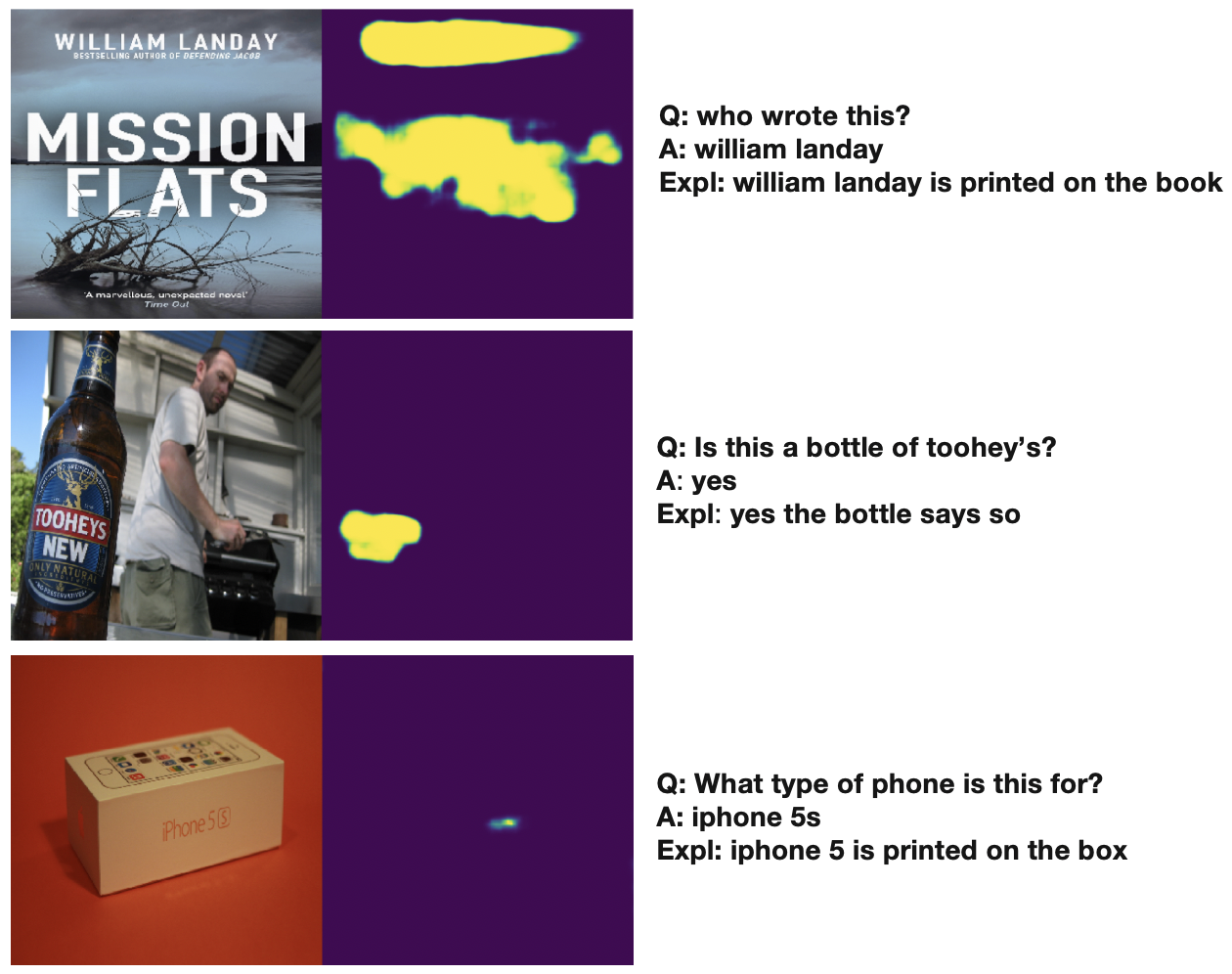}
	\caption{Examples where the MTXNet model produces high quality explanations.}
	\label{fig:successes}
	\vspace{-1.25em}
\end{figure}

\begin{figure}[htb]
	\centering
	\includegraphics[width=\linewidth]{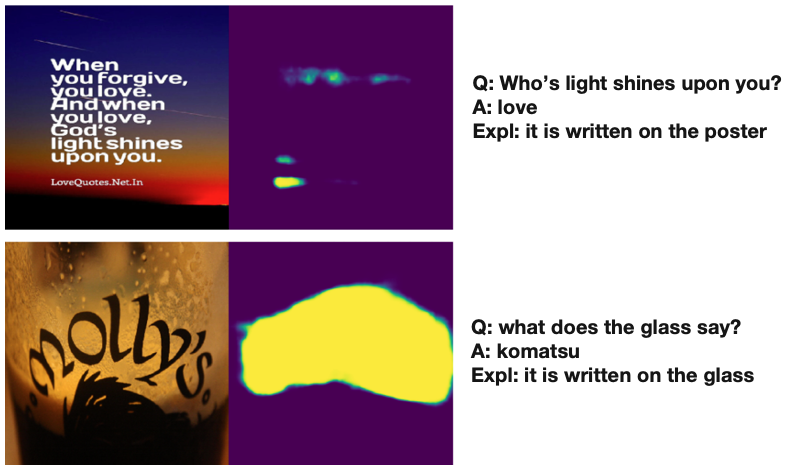}
	\caption{Examples where the MTXNet model fails.}
	\label{fig:failures}
	\vspace{-1.25em}
\end{figure}

\begin{figure}[htb]
	\centering
	\includegraphics[width=0.9\linewidth]{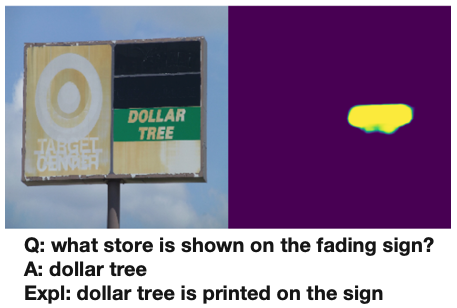}
	\caption{Example where the explanation is consistent with an incorrect prediction.}
	\label{fig:consistentincorrect}
	\vspace{-1.25em}
\end{figure}

As can be seen in Figure \ref{fig:successes}, the MTXNet is able to accurately answer the given question while also justifying its decision through textual and visual explanations. In certain cases, the OCR engine could be inaccurate and lead to wrong tokens being predicted, but the overall answer and explanations are correct. Figure \ref{fig:failures} depicts two failure cases. The upper subimage indicates this could be due to incorrect visual localization while the lower subimage indicates a potential OCR prediction error, although the visual explanation is correct. Despite being generic and dull the textual explanations are correct. In other cases, the model fails due to incorrect visual localization as seen in Figure \ref{fig:consistentincorrect}.

\noindent{\textbf{Explanations help explain incorrect decisions of model.}
	In Figure \ref{fig:consistentincorrect}, we see that the right answer to the question is ``target''. However, the model predicts ``dollar tree''. From the visual and textual explanations we see that the image region localized is incorrect and the model fails to grasp the meaning of ``fading''. This potentially results in it focusing on the more prominent ``dollar tree'' text. Such an analysis provides insights into the component of the system that is failing and deserves further attention.
	
%	\noindent{\textbf{Comparison with VQA-X explanations.} Although we were unable to compare our results qualitatively with that of \citet{huk2018multimodal} \footnote{\url{https://github.com/Seth-Park/MultimodalExplanations}} due to (1) different datasets (2) absence of publicly available inference code and (3) difference in implementation framework (Caffe vs PyTorch), we note that similar to ours, the model proposed in \cite{huk2018multimodal} is also able to generate high quality explanations on generic images and justify incorrect predictions. However, it lacks the ability to to read text in the image and include it as part of the textual explanations. 
		
		% TODO GAT qualitative examples.
		% TODO inference code for mulitmodal explanations not available publicly, cannot compare.
		
		% qualitative results of inference of multimodal explanations paper where text explanations fail.
		
\section{Applications to E-Commerce Businesses} \label{sec:application}
%% talk about explainability, no magic, layman this is how model reasons
%% add in little of this in methods section too.
E-commerce businesses need to comply with industry-wide, and country-specific regulations, to provide accurate and useful information of products to improve customer experience that leads to more business. Our long-term goal with explainable multimodal architectures is to automate and reduce manual effort required for compliance and product detail checks. This will enable businesses to scale compliance and customer experience improvement efficiently without linear increases in cost. Further, these architectures help validate if models are performing as intended and used for the right purposes.

A potential customer experience issue arises when the physical product in a warehouse is different from that uploaded by a seller on the product details page. A possible reason could be that the seller or manufacturer labeled the product erroneously when they packaged it. Many sellers taking advantage of lower cost of manufacturing in a global supply chain, may not be able to audit every batch of product leaving the factory. Such discrepancies will almost certainly lead to product returns, because the customer didn't get what they wanted and increases costs. Such discrepancies may also be due to more nefarious reasons, such as opportunistic bad actors taking advantage of sellers that have successful products by introducing poorer quality or mismatched offers at a lower price to unsuspecting customers. Examples of compliance issues include detecting products that contain batteries and chemicals to comply with transportation and logistics regulations, as well as identifying products that require additional safety documentation and checks, such as products that may have unintended use by children (e.g. toys and products that may end up as toys should not have heavy metals or other poisons that cause illness or death when accidentally ingested). While not all answers can be obtained with product images alone, manual investigation processes utilize these images to identify potential risks that warrant additional steps in the process (e.g. lab testing).

Rather than manually auditing products in a warehouse, product images can be automatically captured at scale, and passed through models that detect such discrepancies. With the help of subject matter experts, attributes such as quantity, color and brand names, and other common misleading attributes are identified apriori. Relevant questions that target these attributes are formulated. The image and question are then inputs to a multimodal explainable system (such as MTXNet) that can provide an answer and justify its prediction through multimodal explanations. Answers can then be compared against the information extracted from the product detail pages on the website. Any discrepancies found can be noted and a selling partner can be provided evidence through the multimodal explanations to take corrective steps. 

An example use-case is as follows. Given a large container of cereal, with smaller boxes within, a potential question is: ``How many cereal boxes are within the container?'' . This information is usually written on the larger container present in the warehouse and can be answered based on reading the text in the image. If there is any discrepancy encountered in the number of boxes of cereal in the warehouse and that listed on the website, appropriate action can be taken. Other similar questions include: ``How heavy is the product?'', ``Is the chair red?'', ``Does the item contain allergens?'', and ``Did the product pass the lead test?''. 

The challenges with the use of such explainable systems are two-fold. First, since there can be multiple stakeholders with diverse expertise and expectations, we need to clearly define the level of abstraction at which they interact with the system. For instance, while a scientist can use the explanations to improve the model, a business operations associate may use the explanations to identify and audit product discrepancies. Second, we need fine grained evaluation methodologies and metrics that take into account the stakeholders as well. \vspace{-0.5em}

 \section{Conclusion}
 \vspace{-0.5em}
A central tenet of explainable AI is to create a suite of tools and frameworks that result in explainable models without sacrificing learning performance and allow humans to understand and trust AI models. As \citet{miller2017explainable} argues, for explainable AI to succeed, we should draw upon existing principles and create strategies that are more people-centric. Unfortunately most prior explanation approaches have been post-hoc, unimodal, ignore text present in the image and not always in accordance with human interpretation. Further, there is a paucity of labeled multimodal explanation datasets. The research presented in this paper shows that existing TextVQA systems can be rather easily adapted to produce multimodal explanations that focus on the text in the image when given access to ground truth annotations. We curate the TextVQA-X dataset consisting of visual and textual explanations. We then present a novel end-to-end trainable architecture, MTXNet, that generates multimodal explanations focusing on the text in the image, in line with human interpretation and surpasses unimodal baselines (7\% in CIDEr scores and 2\% in IoU) while complimenting model performance. We also show how the system may be applicable in the e-commerce space to reduce effort for manual audit of compliance checks and improve customer experience. Results of this research open the door to design of explanainable models part of the original system design that effectively takes advantage of available ground truth multimodal explanation annotations. Future work involves incorporating visual features as part of the transformer architecture.

\bibliography{custom}
\bibliographystyle{acl_natbib}

\end{document}